%% file: main.tex
\documentclass[letterpaper, 10 pt, conference]{ieeeconf}  

\IEEEoverridecommandlockouts                              

\usepackage[sorting=none, style=numeric-comp, maxbibnames=99]{biblatex}

\input{packages}

\input{macros}

\title{\LARGE \bf
Pick2Place: Task-aware 6DoF Grasp Estimation \\via Object-Centric Perspective Affordance 
}

\author{
\authorblockN{Zhanpeng He$^{1,2*}$, Nikhil Chavan-Dafle$^{2}$, Jinwook Huh$^{2}$, %
Shuran Song$^{1,2}$, Volkan Isler$^{2}$\\
$^{1}$Columbia University $^{2}$Samsung AI Center, New York, NY
}
\thanks{$*$ The work was performed when the author was an intern at Samsung AI Center, New York.}
}
\begin{document}

\maketitle
\thispagestyle{empty}
\pagestyle{empty}

\begin{abstract}
The choice of a grasp plays a critical role in the success of downstream manipulation tasks. Consider a task of placing an object in a cluttered scene; the majority of possible grasps may not be suitable for the desired placement.
In this paper, we study the synergy between the picking and placing of an object in a cluttered scene to develop an algorithm for task-aware grasp estimation. 
We present an object-centric action space that encodes the relationship between the geometry of the placement scene and the object to be placed in order to provide placement affordance maps directly from perspective views of the placement scene. This action space enables the computation of a one-to-one mapping between the placement and picking actions allowing the robot to generate a diverse set of pick-and-place proposals and to optimize for a grasp under other task constraints such as robot kinematics and collision avoidance. With experiments both in simulation and on a real robot we demonstrate that with our method, the robot is able to successfully complete the task of placement-aware grasping with over 89\% accuracy in such a way that generalizes to novel objects and scenes.
\end{abstract}

\input{text/intro}
\input{text/related}
\input{text/method}
\input{text/experiments}

\input{text/conclusion}


\printbibliography

\addtolength{\textheight}{-12cm}   
\end{document}

%% file: packages.tex
\usepackage{color}
\usepackage{epsfig}
\usepackage{graphicx}
\usepackage{algorithm,algorithmic}

\usepackage{adjustbox}
\usepackage{array}
\usepackage{booktabs}
\usepackage{colortbl}
\usepackage{float,wrapfig}
\usepackage{framed}
\usepackage{hhline}
\usepackage{multirow}
\usepackage{subcaption} 
\usepackage[font=small]{caption}
\usepackage[percent]{overpic}

\usepackage{amsmath,amsfonts,amssymb}
\usepackage{amsthm} 
\usepackage{bm}
\usepackage{nicefrac}
\usepackage{microtype}
\usepackage{contour}
\usepackage{courier}

\usepackage{changepage}
\usepackage{extramarks}
\usepackage{fancyhdr}
\usepackage{lastpage}
\usepackage{setspace}
\usepackage{soul}
\usepackage{xspace}
\usepackage{cuted}
\usepackage{fancybox}
\usepackage{afterpage}

\usepackage[breaklinks=true,colorlinks,backref=True]{hyperref}
\hypersetup{colorlinks,linkcolor={black},citecolor={green},urlcolor={magenta}}
\usepackage{url}
\usepackage{quoting}
\usepackage{epigraph}

\usepackage{enumerate}
\usepackage{paralist,tabularx}
\usepackage{comment}
\usepackage{pdfpages}
\usepackage{caption}
\usepackage{subcaption}

\usepackage{pifont}

%% file: macros.tex
\usepackage{enumitem}

\makeatletter
\DeclareRobustCommand\onedot{\futurelet\@let@token\@onedot}
\def\@onedot{\ifx\@let@token.\else.\null\fi\xspace}

\def\etal{et al\onedot}

\makeatother

\definecolor{MyDarkBlue}{rgb}{0,0.08,1}
\definecolor{MyDarkGreen}{rgb}{0.02,0.6,0.02}
\definecolor{MyDarkRed}{rgb}{0.8,0.02,0.02}
\definecolor{MyDarkOrange}{rgb}{0.40,0.2,0.02}
\definecolor{MyPurple}{RGB}{111,0,255}
\definecolor{MyRed}{rgb}{1.0,0.0,0.0}
\definecolor{MyGold}{rgb}{0.75,0.6,0.12}
\definecolor{MyDarkgray}{rgb}{0.66, 0.66, 0.66}

\def\OURS{Pick2Place\xspace}

\newcommand{\secref}[1]{Section~\ref{#1}}
\newcommand{\tabref}[1]{Table~\ref{#1}}

\newcommand{\figref}[1]{Fig.~\ref{#1}}
\newcommand{\myparagraph}[1]{\vspace{0.05in}\noindent\textbf{#1}}

\DeclareMathOperator*{\argmax}{argmax} 

%% file: text/intro.tex
\section{INTRODUCTION}
\label{sec:into}

Picking and placing are two fundamental skills that enable diverse robotic manipulation tasks. Many researchers have explored these two tasks independently. 
%
For example, various approaches have been explored to generate six Degrees of Freedom (DoF) grasps that could reliably pick up an object without considering the end-task \cite{Salisbury1982, Bicchi2000, Miller2004, Pinto16grasp, mahler17dexnet2, sundermeyer2021contact, song2020grasping}. A few have studied how to stably place an already grasped object conditioned on the geometries of the object and environment ~\cite{Jiang2012, Holladay2013}.  
This separation is convenient for researchers to reduce the action search space and build robust algorithms. 
However, estimating grasps without considering the downstream task can result in unsuitable grasps for the task. For instance, most of the possible grasps on an object may not be fitting for the task of placing that object in a cluttered environment. 
To make the joint pick-and-place task tractable, many recent works use constrained action space (2D top-down actions) or expensive supervision (expert demonstration on every task) -- limiting their application in novel scenarios requiring high-dimensional action spaces.

\begin{figure}
    \centering
    \includegraphics[width=0.98\linewidth]{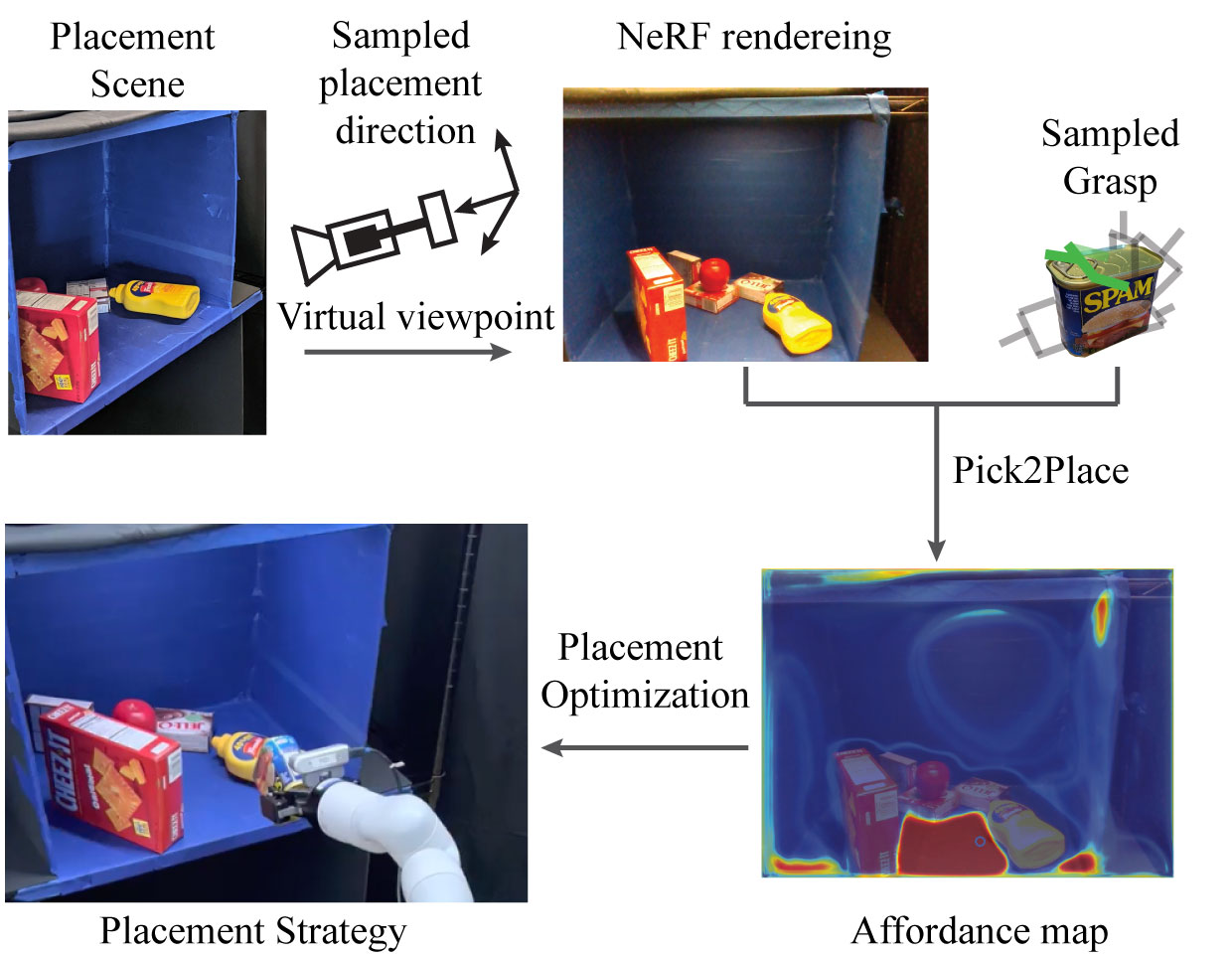}
     \caption{\textbf{Placement-aware grasp estimation}. An example of a placement affordance map and consequent placement strategy generated for a sampled grasp and placement direction. Our method uses a dense set of grasp proposals and placement directions to optimize for the best placement-conditioned grasp estimation.}
    \vspace{-1.5em}
    \label{fig:title}
\end{figure}

In this paper, we present an algorithm, which we call \emph{\OURS}, that addresses the dependency between grasping and placing to perform placement-aware grasp estimation. 
We propose an object-centric action space parameterized by the object transformation required for placement and the placement direction. We show that this action space allows us to establish the correspondence between the picking grasp and the placement pose, naturally providing a method to guide the grasp selection for desired placement and vice versa. 
%

To learn from this action space, we present \textit{object-centric perspective spatial affordance maps}, which provide spatial alignments between actions and observations. Specifically, they capture the affordance for object placement by cross-correlating the encoded geometries of the object and the placement scene with \OURS shown in \figref{fig:method}. 
We integrate our previously developed object reconstruction method~\cite{ChavanDafle2022} and Neural Radiance Field (NeRF) representation~\cite{mueller2022instant} of the scene to generate faithful geometric perspective views of the object and the placement scene. 
Next, we sample over the object transformations and placement directions to optimize for placement-aware grasping as depicted in~\figref{fig:title}.

The integration of NeRF has several advantages: First, it provides high-quality view synthesis; Second, it can accurately recover the geometry of the placing scene which is crucial for reasoning spatial affordance values. Finally, compared to direct rendering \cite{Song2020GraspingIT}, NeRF is robust in reconstructing the geometry of the scene even when commercial depth sensors fail to do so (e.g. transparent \cite{IchnowskiAvigal2021DexNeRF} or shiny objects \cite{YenChen2022NeRFSupervisionLD}). 

We evaluate the performance of our method in a simulation platform as well as on a real robot system. 
We demonstrate that our method is able to capture the influence of the chosen grasp on the placement affordance in the scene. Our results indicate that generating a set of diverse solutions with different object transformations and placement directions allows the robot to optimize the grasping and placement strategy under the constraints imposed by the robot's kinematics and scene geometry. The robot is able to complete the object insertion and placement in shelf tasks, as shown in \figref{fig:tasks}, with about 89\% accuracy and outperforms the baseline methods.  

We summarize our key contributions as follows:
\begin{compactitem}
    \item We provide an object-centric action space that correlates the geometry of an object to that of the manipulation scene for the 6DoF pick and place task. 
    %
    %
    \item We integrate Neural Radiance Fields (NeRF) into spatial action map learning that allows for generating placement affordance maps conditioned on the object grasp and placement directions.
    
    %
    %
    \item Our experimental results, both in simulation and on a real robot setup, show that a robot can compute and successfully execute task-aware grasps using our formulation.
\end{compactitem}

With our work on placement-aware grasp planning, robots can not only grasp but also use objects for a desired task. Such object-rearrangment skill is essential as we envision robots performing day-to-day tasks in unstructured settings.

%% file: text/related.tex
\section{Related Work}
\begin{figure*}[t]
    \centering
    \vspace{0.1em}
    \includegraphics[scale=0.72]{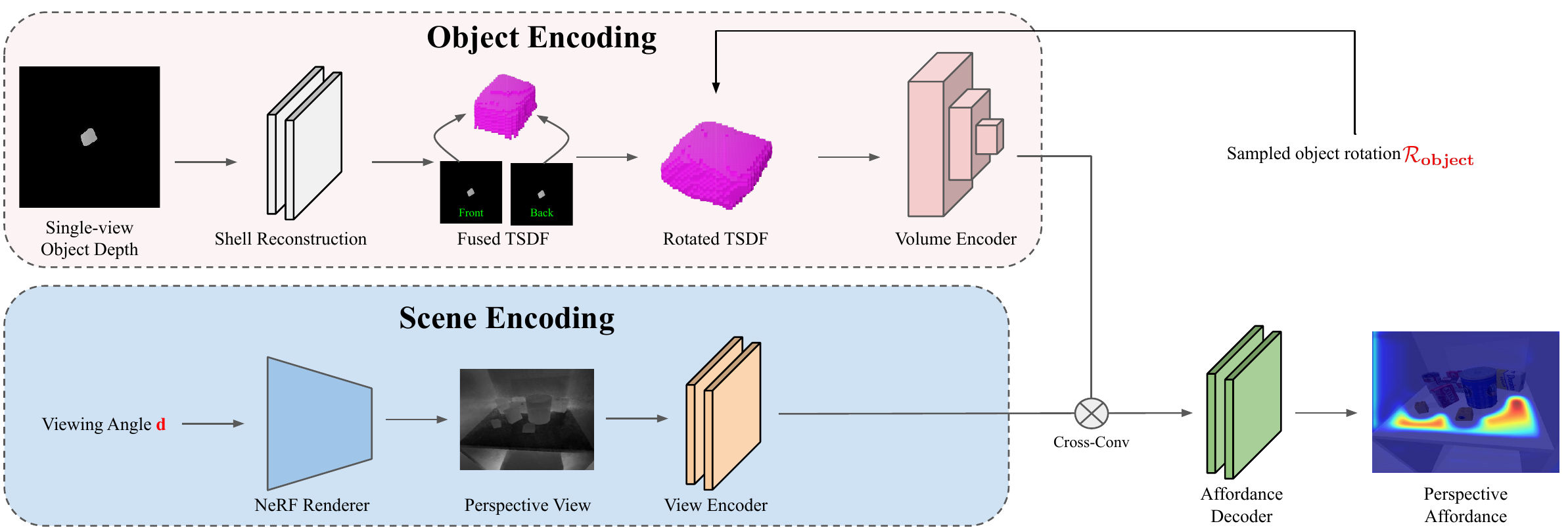}
    \caption{\textbf{Approach overview}. To evaluate a placement of an object with rotation $\mathbf{\mathcal{R}_{object}}$, we cross-correlate the encoded object with the encoded scene to perform pattern matching. \OURS uses NeRF as scene representation. It renders the scene from various viewing directions $\textbf{d}$ and computes spatial action values for each of the views. To encode an object, we use the TSDF of the object shell reconstruction~\cite{ChavanDafle2022} from a single view depth image.}
    \vspace{-1.5em}
    \label{fig:method}
\end{figure*}
\label{sec:related}

Object rearrangement, i.e., moving an object from one pose to another is a crucial robotic manipulation skill. Being able to grasp an object is only the part of the process, but often dictates the successful execution of the task. 

\subsection{Grasp planning}
\label{sec:grasp_plan}
Grasp planning has been an active topic for robotic manipulation research and applications for the last few decades. 
Early work focused on developing grippers and analytical planning methods to grasp a wide range of objects and for precision-based manipulation~\cite{Salisbury1982, Bicchi2000, Roa14graspq, Miller2004}. 
Many of these early work assumed the knowledge of the object geometry and object pose to compute grasps. 
Recent learning based methods directly act on depth images \cite{ChavanDafle2022} or pointclouds~\cite{Pas2017GraspPD, mahler17dexnet2,sundermeyer2021contact} of the scene to generate grasp proposals on novel objects in the scene. 
Most of these works however consider the problem of grasp planning with little consideration to the intended use of the grasped object. 
%
In many applications the object removal is not sufficient and being able to place the grasped object at the designated location is more important. 

\subsection{Object Placement}
\label{sec:placement}
A few researchers have studied the problem of object placement. Some of the early works look into task-level planning for pick and place tasks and adding more primitives such as pushing for table-top placements~\cite{LozanoPerez1989, Sugie1995}. 
%
To consider the stability of placements, Fu \etal~\cite{Fu2010} and Saxena \etal~\cite{Saxena2009} focused on defining the features that allow a robot to learn upright orientation of the objects and stable placements on a flat surface respectively. These approaches assume a known 3D model of the object or require a complete pointcloud of the object. 
Schuster \etal developed a method to find flat empty areas in the scene from an image and applied it to object placement~\cite{Schuster2010}. However, they do not consider the dependence of object orientation or the placement direction on the placement affordance, which we study in this paper.

\subsection{Task Aware Grasping and Manipulation}
\label{sec:post_grasp}

With increasing application of robotic manipulation in logistics and manufacturing automation and potential for robots in home, there is more research on complete task level solutions and systems. Many of the recent works study putting together components, such as perception, grasp planning, robot motion planning, task planning, etc., which often are more extensively studied in isolation. 
For example, Wang \etal ~\cite{Wang2019} study the combined robot motion and grasp planning which allows the robot to select the grasp reachable for the robot. Wada \etal \cite{Wada2022} develop a framework to reorient the object to achieve the desired placement in the shelf. They assume known objects with 3D models and goal pose specification to define the task. Zeng \etal \cite{zeng2020transporter} present a neural network to predict pick-conditioned placing actions. Our \OURS method generalizes to novel objects and outperforms \cite{zeng2020transporter} for object insertion
and placement tasks.

\subsection{Neural Radiance Fields}
Researchers have shown impressive results on view synthesis~\cite{Mildenhall2020NeRFRS, Deng2021DepthsupervisedNF, Yu2021pixelNeRFNR} with NeRF and using NeRF for representing scenes \cite{Pumarola2021DNeRFNR}. 
Several works have explored using NeRF for robotic tasks. Lin \etal \cite{YenChen2022NeRFSupervisionLD} investigate using NeRF to synthesize views of an object for learning dense object descriptors for grasping. Ichnowski \etal~\cite{IchnowskiAvigal2021DexNeRF} use NeRF to improve the estimation of geometry of transparent objects and thus improve the grasping success. While these works use NeRF for picking, \OURS focuses on applying NeRF for placement prediction. 
Many methods for optimizing NeRF suffer from long training time, and hence NeRF is difficult to be used in robotic tasks. A recent work \cite{mueller2022instant}, named Instant-NGP, shows that NeRF can be trained in 5 seconds and still preserve robust rendering performance with a multi-resolution hash encoding. In our work, we use Instant-NGP to train NeRF and render images for affordance estimation.

%% file: text/method.tex
\section{Method}
\label{sec:method}

\subsection{Problem Formulation}
\label{sec:probelm-formulation}
We formulate the problem of rearranging objects as learning actions $a_t$ from observations $o_t$:
\[
\pi(o_t) \rightarrow a_t \in \mathcal{A}.
\]

As shown in a previous work \cite{zeng2020transporter}, $\mathcal{A}$ can be parameterized by $\{\mathcal{T}_{pick}, \mathcal{T}_{place}\}$ , where $\mathcal{T}_{pick}$ and $\mathcal{T}_{place}$ represent the pose of the end-effector when grasping the object and when releasing the grasp respectively. 
Parameterizing the pick and place actions as two poses of the end-effector allows for designing efficient algorithms to learn spatial affordance maps in 3D space. 
$\mathcal{T}_{pick}$ and $\mathcal{T}_{place}$ are related since both actions together can be used to decide $\mathcal{T}_{object}$, the object transformation for placement. Hence, in order to learn the full pick and place task, the model needs to learn the inherent pattern matching between the object geometry and the placement scene from the poses of the end-effector.

In this work, we leverage the synergies between pick and place and explicitly parameterize the action space in an object-centric manner:
\vspace{-3mm}
\[
a_t = \{\mathcal{T}_{pick}, \mathcal{T}_{object}, a_{insert}\},
\]
where $\mathcal{T}_{pick}$ is the pose of the end-effector when grasping, $\mathcal{T}_{object}$ represents the object transformation when placing~\footnote{Note that $\mathcal{T}_{object}$ is the transformation of the object compared to its initial configuration before picking. Also, it is not limited to object pose representation only. For example, in our implementation, $\mathcal{T}_{object}$ is the transformation of the object TSDF computed from the initial view.}, and $a_{insert}$ represents the direction of a translation action of the end-effector to reach $\mathcal{T}_{object}$. In fact, with this object-centric parameterization, if we can compute two of the three actions, we can infer the rest by making the assumption that the end-effector normal direction (gripper palm normal) aligns with $a_{insert}$:
\vspace{-1mm}
\[
\mathcal{T}_{pick} = f_{p}(\mathcal{T}_{object}, a_{insert}),
\]
where $f_{p}$ represents a function that maps a placement action to a picking action. 
One way to parameterize $f_{p}$ is using a neural network. However, if $f_p$ is learned from data, the relation between pick and place can only be enforced by expert demonstrations successfully completing the task. Furthermore, neural networks can give wrong estimations if the input data are out of the training distribution and hence can lead to infeasible pick actions.
In our work, we directly implement this function and do not learn it from data. We will discuss the conversion between placing and picking actions as well as estimating placement performance of a grasp in~\secref{sec:synergy-pick-n-place}.


\subsection{Learning perspective affordances for placing}
\label{sec:learning_afforance}
In this work, we aim to learn SE(3) poses for both pick and place actions. This is challenging because of the high-dimensional action space. While many works have shown promising results in learning spatial action maps for SE(2) action space, applying spatial action maps in SE(3) action space is not straightforward due to the difficulty of aligning 3D spatial information with the action space.
To address this problem, we propose to learn object-centric perspective spatial affordance maps. Concretely, our goal is to produce affordance score maps $s_{a_{insert}, R_{object}}$, whose pixels represent the score of a placing action with specific insert direction $a_{insert}$ and a specific object transformation $\mathcal{T}_{object}$.  

To preserve 3D spatial information of the placement scene and the object to be placed during spatial affordance map learning, we integrate NeRF \cite{mueller2022instant} and object shell reconstruction~\cite{ChavanDafle2022} respectively. We use NeRF as a scene representation and also a neural renderer to provide perspective information that aligns with the action space. Specifically, we represent each placement scene as a NeRF by optimizing a Depth-Supervised NeRF (DS-NeRF) model \cite{Deng2021DepthsupervisedNF}. To compute the spatial alignment, when evaluating the affordance value of insert direction $a_{insert}$,  we render a depth image using the optimized NeRF model from viewing direction $\textbf{d} = a_{insert}$ and encode it with an image encoder $p$. 
%
%
Then, a placing action can be seen as a pattern-matching problem between the object geometry and the rendered local 3D geometry of the scene. The object shell reconstruction allows us to predict the 3D geometry of the object from a single depth image. We use truncated signed distance function (TSDF) of the object reconstruction to represent an object that is being placed. To evaluate across orientations of the object, we rotate the TSDF with a specific rotation $R_{object}$. Then, the rotated TSDF is encoded using an object encoder $q$ to produce kernels that will be used to cross-correlate object and scene information. 

In summary, as shown in Figure \ref{fig:method}, our place model $f_v$ is an affordance value function that is composed of three components. First, a NeRF encapsulates geometric information from a clutter placement scene and is used for rendering perspective images providing local geometry; then, a view encoder encodes the perspective view. Second, to integrate the object's geometric information for placing, an object is represented using TSDF and encoded by an object encoder to produce image kernels. Finally, to evaluate a placement, we cross-correlate the encoded object and encoded scene to produce an affordance map. To derive an optimized placement, we sample different actions ($R_{object}$ and $a_{insert}$ tuples), feedforward them through the network to produce a set of affordance maps and take the argmax as:
\vspace{-1mm}
\begin{align}
\centering
\pi(o_t) &= a_t = \argmax_{a_{place}} f_v(o_t, a_{place})  \nonumber \\
\text{where } &a_{place} = \{u, v, R_{object}, a_{insert}\}. \nonumber
\label{eq:lower-bound}
\end{align}
Here, $u, v$ are the pixel location on the affordance map which map to the placement position.

\subsection{Conversion between pick and place}
\label{sec:synergy-pick-n-place}

After the successful execution of a grasp, the object is rigidly attached to the end-effector of the robot. 
In this work, we impose constraints that during the placement action the palm normal of the gripper aligns with the insertion direction $a_{insert}$. Moreover, the gripper is orientated such that the camera is on the upper side (towards the world frame $Z$ axis). Therefore, a sampled insertion direction completely defines the orientation of the gripper during placement action. Given a grasp action $\mathcal{T}_{pick}$ and a sampled insertion direction $a_{insert}$, the required object rotation $\mathcal{R}_{object}$ (same as the end-effector rotation) for the placement action is given as $\mathcal{R}_{object} = \mathcal{R}_{pick}^{-1}\mathcal{R}_{insert}$. Here, $\mathcal{R}_{pick}$ and $\mathcal{R}_{insert}$ are the SO(3) orientations of the end effector when picking and placing the object, respectively.


We exploit our prior work on simultaneous shape reconstruction and grasp planning~\cite{ChavanDafle2022} to generate a dense set of feasible grasps on the object from a depth image of the object.
For the set of sampled grasps and insertion directions, we generate affordance maps and take the argmax to compute the placement action, specifically, the placement location, as explained in \secref{sec:learning_afforance}.
The placing position is given by converting a pixel location on the image to the world coordinate. To avoid collision of the object with the placement surface, we compute an offset on the World Frame $Z$ axis using our reconstructed TSDF for placing.

\subsection{Implementation details}
\myparagraph{Object representation and encoding:} The object geometry is captured by reconstruction from a single view depth image. We use an in-house shell reconstruction network~\cite{ChavanDafle2022} to predict the exit depth image and encode the object as a 3D TSDF volume. Then the volume encoder is 4 layers of 3D convolution layers that reduce the z dimension to 1, followed by four 2D convolution layers that produce object features.

\myparagraph{Scene encoding and decoding:} The rendered perspective depth image of shape $H \times W \times 1$ from NeRF is encoded by the view encoder with 5 layers of 2D convolution with a kernel size of 3. Then, to perform pattern matching between the placement scene view and the object to be placed, we apply convolution using the object features as kernels. Finally, the output of the cross-convolution is used to decode an affordance map in shape $H \times W \times 1$.

\myparagraph{Loss function and training:} Finally, the model is trained as a binary classification with a cross-entropy loss $L_{pix}$. During training on the shelf-placing task, for learning via trial-and-error, we sample $16$ perspective views and $72$ object orientations for each episode, and employ epsilon greedy as an exploration strategy to produce actions for data collection.

%% file: text/experiments.tex
\begin{figure}[h]
\vspace{-1.em}
    \centering
    \includegraphics[width=0.98\linewidth]{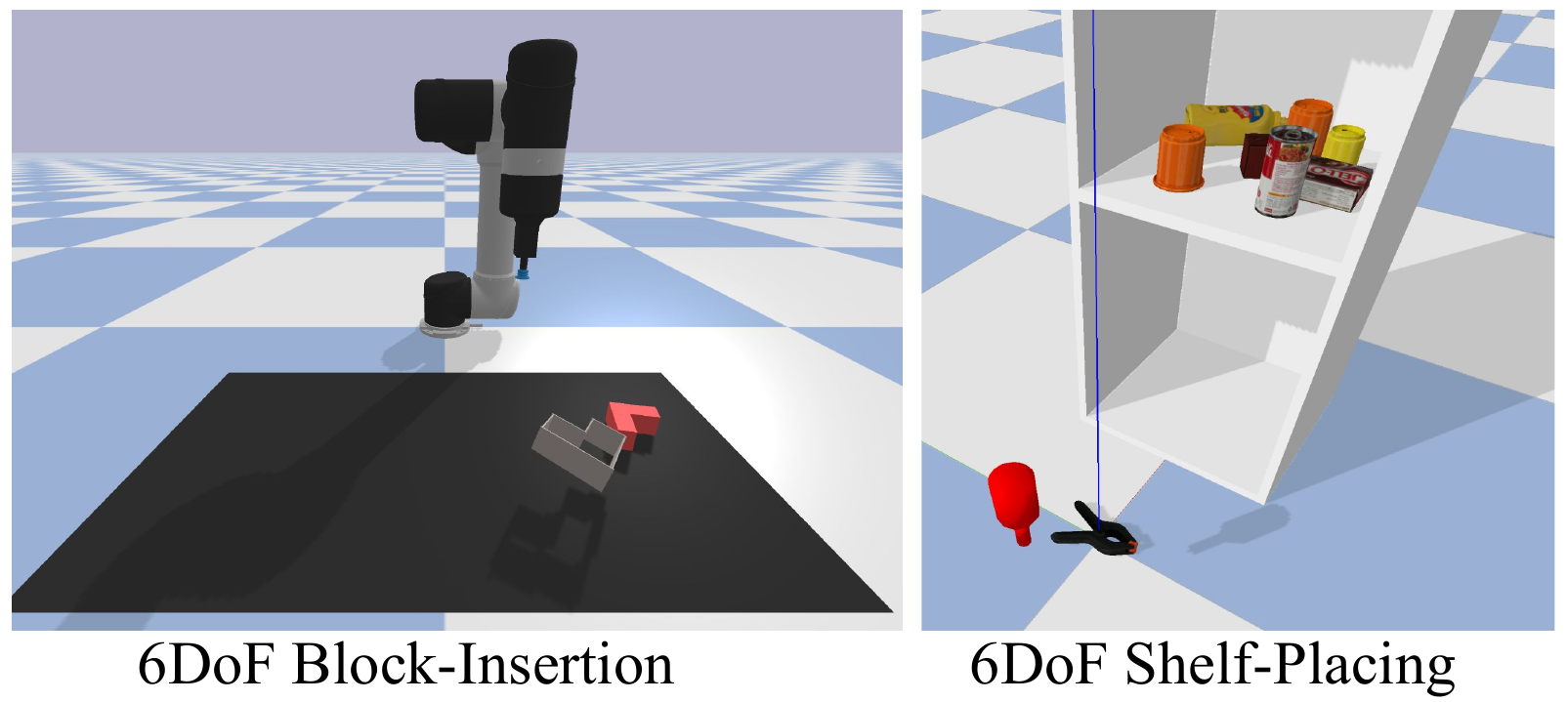}
    \caption{\textbf{Visualization of evaluation tasks.} Left: 6DoF block-insertion; Right: 6DoF shelf-placing.}
    \vspace{-1.5em}
    \label{fig:tasks}
\end{figure}
\section{Experiments}
\label{sec:experiments}

We demonstrate that \OURS can be used for different pick and place tasks. Picking considering placement conditions is particularly important when the task is kinematically constrained such as placing the object in tight spaces such as shelves or placing the objects precisely in fixtures since many of the grasps are not suitable for the task. 
In our experiments, we first verify that our method can be used for behavioral cloning on the benchmark task proposed by the Transporter Network \cite{zeng2020transporter}. Then, we evaluate the performance of \OURS with pick and place tasks in a shelf environment both in simulation and on a real robot setup.
\begin{figure*}[t]
    \centering
    \vspace{0.5em}
    \includegraphics[width=0.97\linewidth]{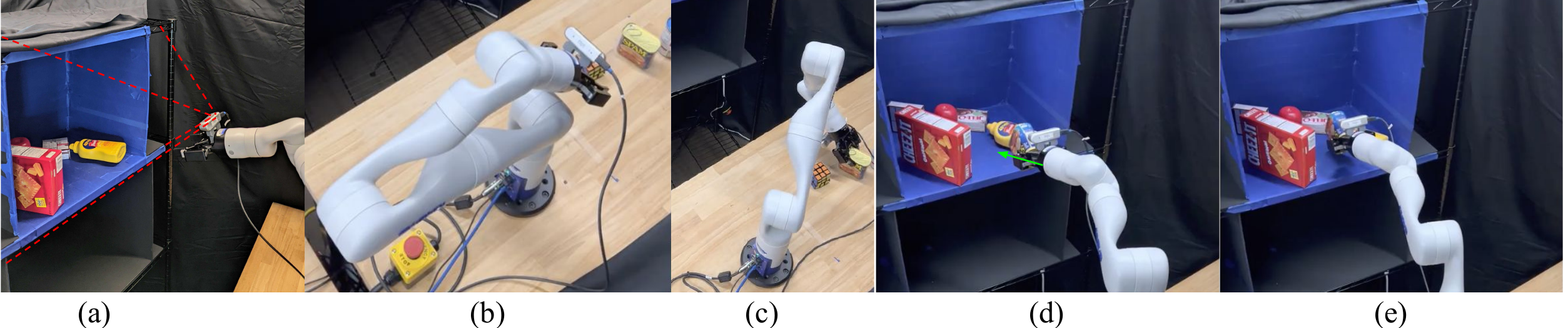}
    \caption{\textbf{Executing pick and place in the real world.} 
    The robot (a) acquires images of and builds the NeRF representation of the scene, (b) acquires an image of the object, completes the object geometry using the Shell reconstruction and iterates over object rotations encoded and cross-correlated with the scene so as to compute the grasp and placement poses (c) the grasp is executed for the picking action, and placement is completed by moving to a pre-placement pose (d) followed by insertion (e).}
    \label{fig:realworld-execute}
    \vspace{-1.3em}
\end{figure*}
\subsection{Baselines}
\label{sec:baselines}
We compare our method with two baselines:  
\begin{compactitem}
    \item \textbf{GT-State MLP:} This method takes the groundtruth state of the objects as inputs and infers two SE(3) poses for pick action and place action using multi-layer perceptrons (MLP). Our method and TransporterNet-based method learn an action value function for a task. GT-state MLP, instead, directly outputs actions without considering their value functions.
    \item \textbf{TransporterNet-SE(3):} This method first infers 6DoF actions by computing SE(2) action from topdown view of the scene in the form of spatial action map, then regress the rest of the action space using MLPs.
\end{compactitem}

\subsection{6DoF block-insertion}
\label{sec:exp_block}

As shown in Figure \ref{fig:tasks}, we first verify our method with a benchmark task introduced in~\cite{zeng2020transporter}: a 6-DoF L-shape block insertion that requires an L-shape object to be placed in an arbitrarily-oriented L-shape fixture. 
This task is challenging since it requires the agent to perform precise matching between the object and the target fixture. Hence, there exists only a few solutions to this task. 

In this experiment, we evaluate \OURS and the baselines in two settings: 1. target place poses are randomly sampled from the training distribution (rotations of the fixture are sampled from: $\theta_x, \theta_y \in [\frac{-\pi}{5}, \frac{\pi}{5}], \theta_z \in [-\pi, \pi]$); 2. target place configurations are out of the training distribution (rotations of the fixture are sampled from: $\theta_x, \theta_y \in [\frac{-2\pi}{5}, -\frac{\pi}{5}] \cup [\frac{\pi}{5}, \frac{2\pi}{5}] , \theta_z \in [-\pi, \pi]$). The first setting checks whether our model is able to generalize to pick-and-place tasks that are similar to the training demonstrations. The latter one challenges our model to find views that can generate feasible solutions. 
\vspace{-0.5em}
\begin{table}[h]
\centering
    \begin{tabular}{ccccccc} 
    \toprule
    & \multicolumn{3}{c}{\shortstack{\small{block-insertion} \\ \small{(ID test poses)}}} & \multicolumn{3}{c}{\shortstack{\small{block-insertion} \\ \small{(OOD test poses)}}}  \\
    & \multicolumn{3}{c}{\rule{2cm}{0.2pt}} & \multicolumn{3}{c}{\rule{2cm}{0.2pt}} \\
    Method & 1 & 10 & 100 & 1 & 10 & 100 \\
    \midrule
    GT-State MLP & 0 & 1 & 1 & 0 & 0 & 0 \\
    TransporterNet-SE(3) & \textbf{28} & 76 & 81 & 0 & 13 & 20  \\
    Ours & 1 & \textbf{85} & \textbf{89} & 0 & \textbf{72} & \textbf{75} \\
    \bottomrule
    \end{tabular}
    \caption{\textbf{Quantitative results for the 6DoF block-insertion task.} Task success rate (mean \%) vs. \# of demonstration episodes (1, 10, or 100) used in training. ID represents in-distribution test poses, and OOD represents out-of-distribution test poses.}
    \vspace{-1em}
    \label{tab:block-insertion result}
\end{table}

As shown in Table \ref{tab:block-insertion result}, GT-State MLP model fails to learn in-distribution poses as well as out-of-distribution poses. TransporterNet-SE(3) learns reasonably well on placing objects in poses within the training distribution, but its performance drops when testing with out-of-distribution cases. A major cause of this performance drop is the difficulty of inferring solutions from the top-down view of OOD target poses. In some extreme cases, a solution may be occluded in the top-down view. Our method is able to learn the block insertion task while generalizing to OOD target place poses. \OURS has a bad performance when only one demonstration is available for training. This can be due to the lack of data variety in the training data.

\subsection{Placing objects to a shelf}
\label{sec:exp_shelf_sim}
\begin{figure}[t]
    \centering
    \includegraphics[width=0.98\linewidth]{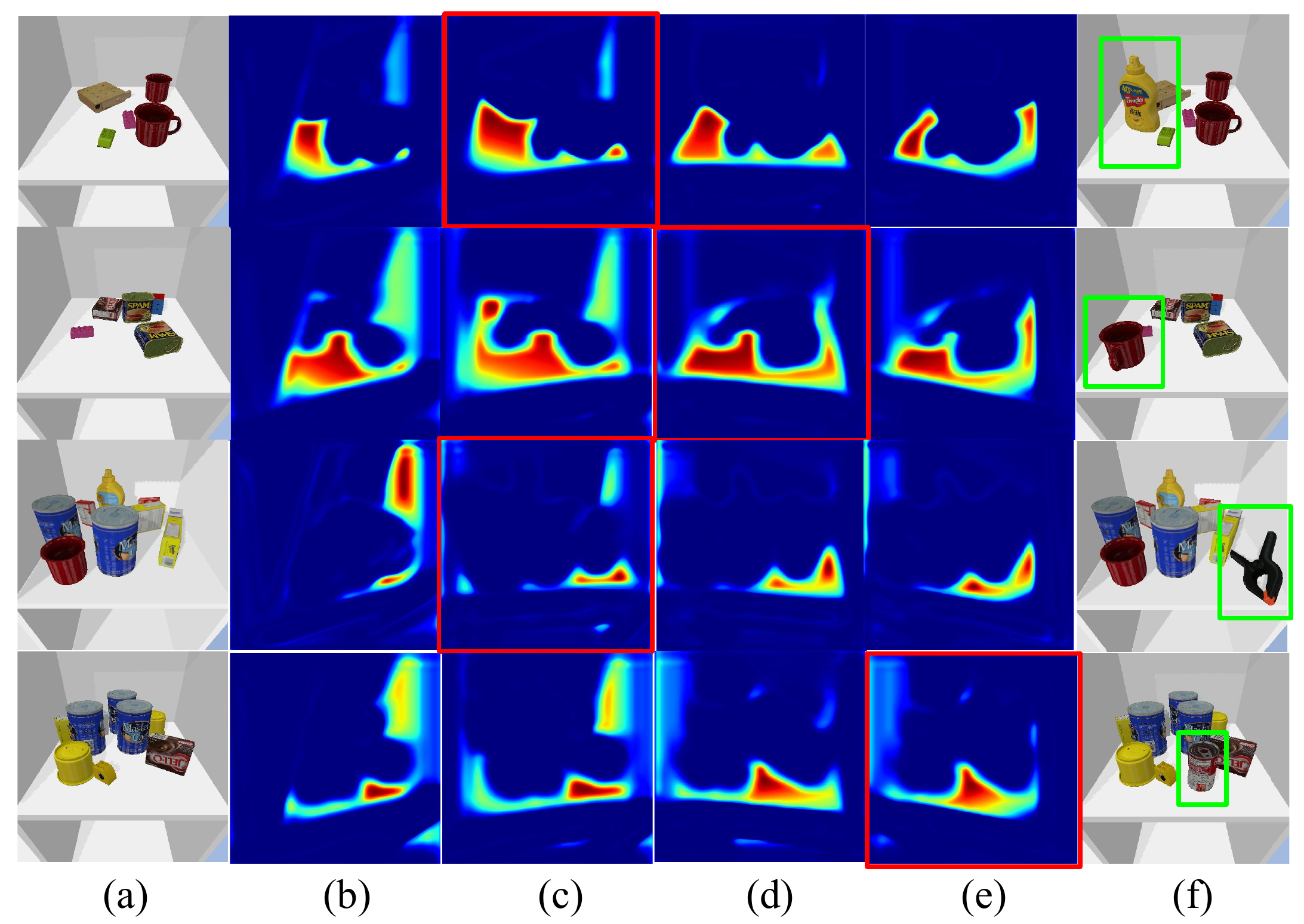}
    \caption{\textbf{Qualitative results for the shelf-placing task.} (a) scene images, (b) - (e) the affordances from different viewing angles, (f) the scene after the object is stably placed (shown in green). The red rectangle highlights the selected view for action execution. The first two rows show examples from the training distribution while the last two rows are test cases with  higher scene complexity.}
    \label{fig:qualitative}
    \vspace{-1.6em}
\end{figure}
In the second task, we test whether our model can be used as an action-value function in a trial-and-error setting. In this task, we ask the agent to use a virtual gripper to pick up an object and place it in the higher section of a shelf. The solutions to this task are not unique and a placement is considered as success if it meets the following requirements: 1. the placing object does not collide with any other object in the scene; 2. the placing object is stably placed within the target region. In this experiment, we do not provide any demonstration during training, instead train a policy to complete the task via trial-and-error using sparse reward signals. The agent can only get a reward of 1 if the object is successfully placed for each episode. 

For each episode, $N$ objects are randomly initialized in the shelf where $N \in [5, 7]$ during training. We train the model with a set of synthetic shapes from ~\cite{ChavanDafle2022}. During testing, we evaluate each placement with higher scene complexity where each scene has 5 to 9 objects from YCB dataset. In this task, to build the NeRF representation of a scene, we sample 24 viewing angles that look at the scene center.

\begin{figure}[t]
    \centering
    \vspace{0.1em}
    \includegraphics[width=\linewidth]{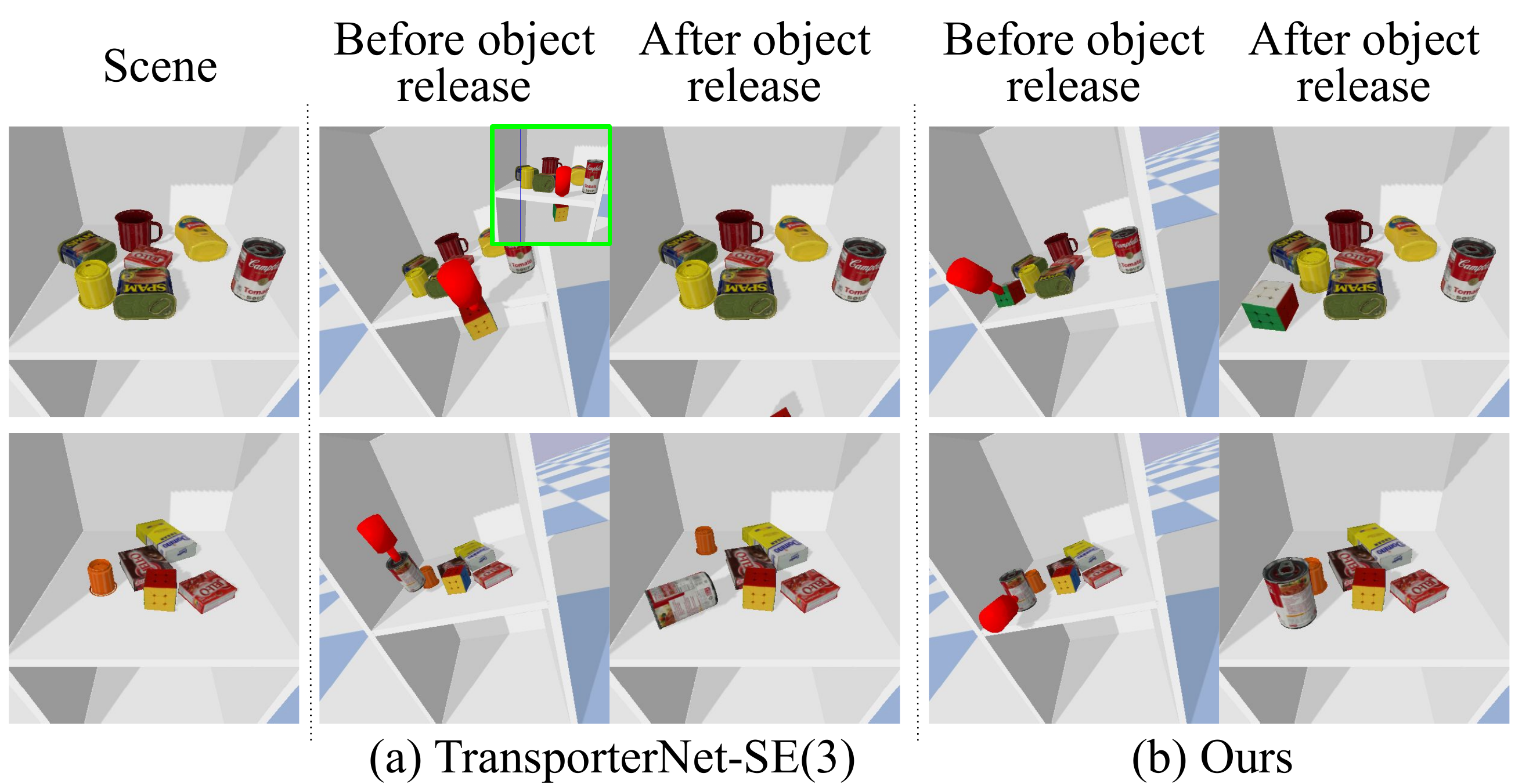}
    \caption{\textbf{Examples of failure cases from TransporterNet-SE(3)~\cite{zeng2020transporter}.} Each row shows a test case. For test case 1, the predicted place action by TransporterNet-SE(3) leads to collision with the shelf before reaching the placing pose, however we render the estimated placing pose in the upper-right green box for clarity.}
    \label{fig:failures}
\end{figure}

\begin{figure}[t]
    \centering
    \includegraphics[width=0.98\linewidth]{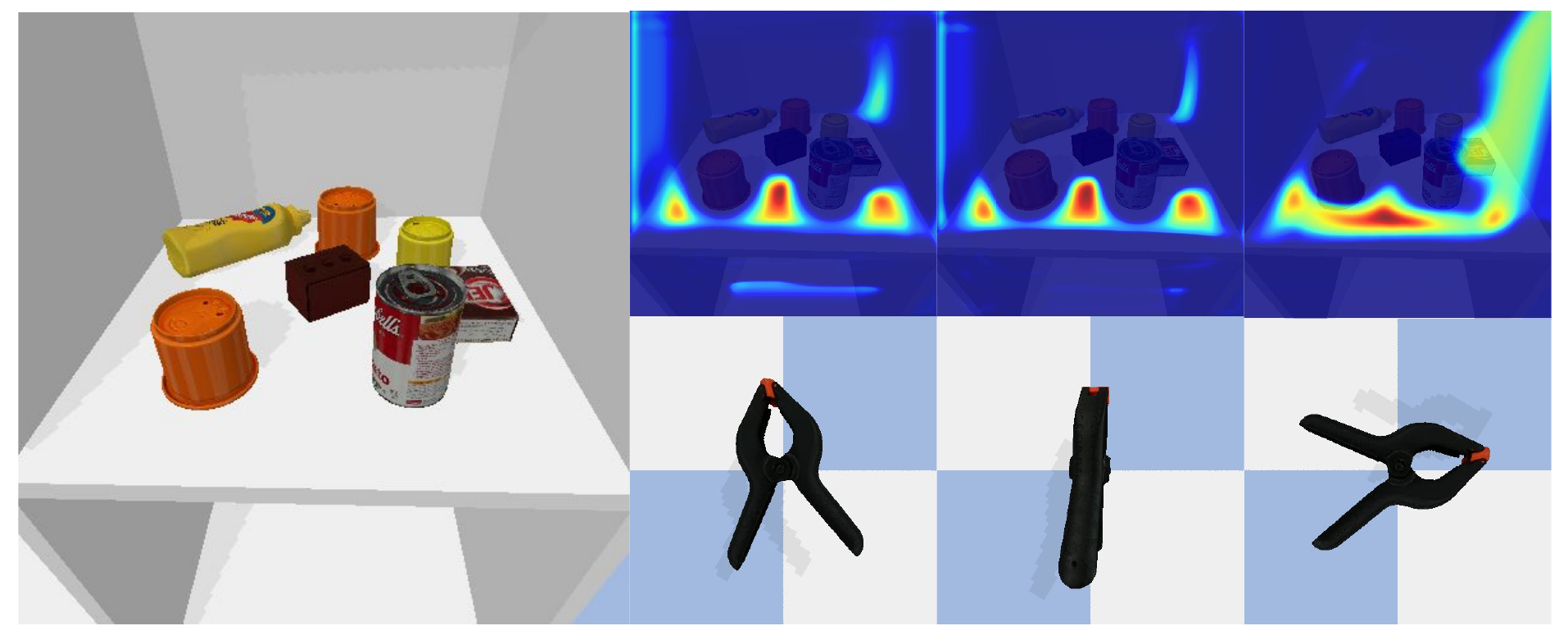}
    \caption{\textbf{Influences of object orientation} Qualitative results of \OURS show that different placement actions are preferred when placing with different object orientations.}
    \label{fig:poses}
    \vspace{-1.5em}
\end{figure}
\begin{table}[h]
\centering
    \begin{tabular}{ccc}
    \toprule
    & \multicolumn{2}{c}{shelf-placing} \\
    \small{Method} & \small{Fully-accessible} & \small{Limited workspace}  \\
    \midrule
    \small{GT-State MLP} & 0 & 0 \\
    \small{TransporterNet-SE(3)} & 80 & 53 \\
    \small{Ours} & \textbf{89} & \textbf{78} \\
    \bottomrule

    \end{tabular}
    \caption{Quantitative results for the 6DoF shelf-placing task. }
    \vspace{-.5em}
    \label{tab:shelf result}
\end{table}

As shown in~\tabref{tab:shelf result}, GT-State MLP fails to learn the task in the sparse reward setting due to its failure to explore any positive reward signals. TransporterNet-SE(3), although reach a reasonably high success rate, its regression module sometimes produces inaccurate placing heights or rotations for test objects. 
For example, as shown in~\figref{fig:failures}-(a), although it infers correct $x$ and $y$ locations for placing on the image space, a wrong $z$ and rotation estimation can result in an unstable placing action that drops the object and causes a collision between the placed object and scene objects. Another failure case is that TransporterNet-SE(3) estimates a low placing height that causes the collision between the object and the shelf. The green box in~\figref{fig:failures}-(a) shows that the pose estimation from TransporterNet-SE(3) and it is about $0.05$m lower than a feasible placement. Finally, \OURS is able to reach the highest success rate in these tasks. As shown in~\figref{fig:qualitative}, our model can discover diverse solutions: with different insertion directions, it generates different feasible regions in the scene to produce collision-free placement. 

We demonstrate that finding diverse solutions is useful when the placing task with kinematic constraints. For example, if only a part of the shelf is within reach of the robot, solutions can still be found in some of the perspective views. To verify this, we restrict the workspace of the virtual gripper to be the volume of a ball with a center of $(0.2,  -0.4, 0.5)$ and a radius of $0.5$. As shown in~\tabref{tab:shelf result}, since TransporterNet-SE(3) always produces a single solution for each scene, it has worse performance when the workspace is limited. Although the performance of our method also gets adversely affected, it still shows a relatively high success rate.

We also show that our model learns to match the geometry of the object with that of the placement scene. For example, when estimating the placement of the clamp, as shown in~\figref{fig:poses}, our model utilizes different regions of the scene to place the object in different orientations.
\begin{figure}[t]
    \centering
    \vspace{0.25em}
    \includegraphics[width=0.98\linewidth]{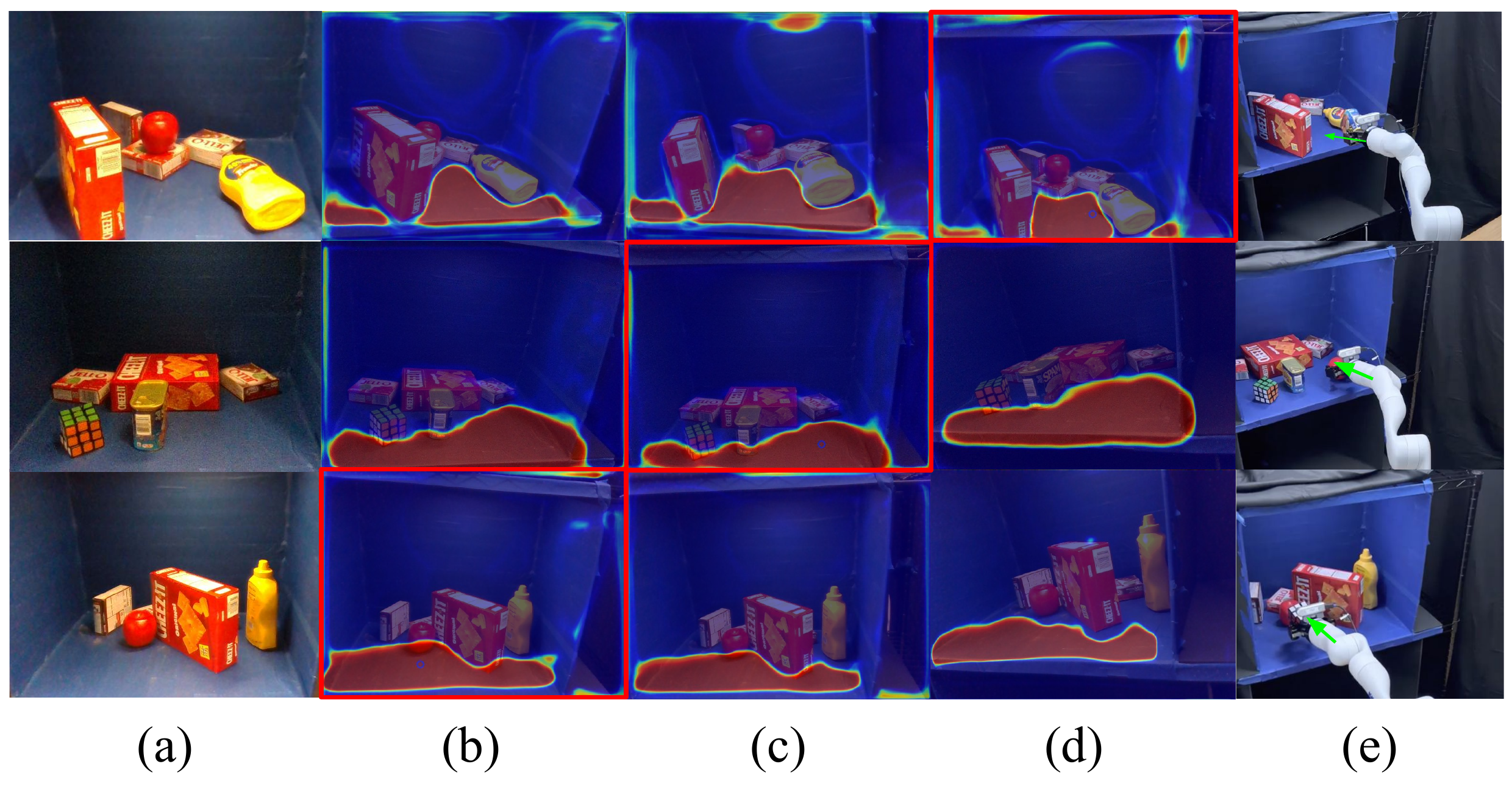}
    \caption{\textbf{Qualitative results for real robot validation.} Column (a) shows the placing scene. Column (b) - (d) show examples of perspective NeRF views, sampled from different insertion direction, overlaid with the corresponding placement affordance maps. Column (e) shows the execution of a placement.}
    \vspace{-1.3em}
    \label{fig:real-qual}
\end{figure}

\subsection{Real robot experiment}
\label{sec:exp_shelf_real}

We validate \OURS for the shelf-placing task in real world using a Kinova robot arm, with a RealSense Camera D435 mounted on its wrist, and a two-finger Robotiq gripper. To execute a placement on the shelf, the robot first moves to 20 locations and captures RGB images of the scene. We use COLMAP \cite{Schonberger2016colmap} to refine the camera poses and intrinsics and derive sparse depth information for NeRF optimization. We build the NeRF representation of the scene and feed forward \OURS with sampled grasps and insertion directions to find the placement with the highest affordance score.  \figref{fig:realworld-execute} shows an example of the execution of the pick and place pipeline. \figref{fig:real-qual} shows the performance of \OURS with a synthesized scene views from real-world data. The supplementary video shows representative executions of picking and placing with our model.

%% file: text/conclusion.tex
\section{Conclusion and Future Directions}
\label{sec:conclusion}

We introduced \OURS, a method to estimate placement-aware grasps via object-centric perspective affordance. 
We proposed an object-centric action space that correlates
the geometry of an object to that of a placement scene for generating the 6DoF pick and place strategies. We demonstrated that \OURS can explore the space of placement directions to compute and optimize for the placement affordance by using NeRF as scene representation and renderer. With experiments both in simulation and on a real robot setup, we validated our proposed method to successfully complete the task of picking up an object and placing it in a cluttered scene.


Although \OURS  requires a time-consuming process for capturing multiple images by a wrist-mounted camera for NeRF representation, this limitation can be easily addressed by using multi-camera systems or algorithms that require only a few images to train NeRF (e.g., PixelNeRF \cite{Yu2021pixelNeRFNR}).
Since our \OURS pipeline is differentiable, there are many exciting future directions. In particular, we plan to incorporate our pipeline into a action planning strategy.

%% file: references.bib
@INPROCEEDINGS{Schuster2010,
  author={Schuster, Martin J. and Okerman, Jason and Nguyen, Hai and Rehg, James M. and Kemp, Charles C.},
  booktitle={2010 10th IEEE-RAS International Conference on Humanoid Robots}, 
  title={Perceiving clutter and surfaces for object placement in indoor environments}, 
  year={2010},
  pages={152-159},
  }

@article{song2020grasping,
title={Grasping in the Wild: Learning 6DoF Closed-Loop Grasping from Low-Cost Demonstrations},
author={ Song, Shuran and Zeng, Andy and Lee, Johnny and Funkhouser, Thomas},
journal={Robotics and Automation Letters},
year={2020} 
}

@article{Jiang2012,
author = {Yun Jiang and Marcus Lim and Changxi Zheng and Ashutosh Saxena},
title ={Learning to place new objects in a scene},
journal = {IJRR},
volume = {31},
number = {9},
pages = {1021-1043},
year = {2012},
}

@INPROCEEDINGS{Sugie1995,
  author={Sugie, H. and Inagaki, Y. and Ono, S. and Aisu, H. and Unemi, T.},
  booktitle={IEEE ICRA}, 
  title={Placing objects with multiple mobile robots-mutual help using intention inference}, 
  year={1995},
  volume={2},
  number={},
  pages={2181-2186 vol.2},}

@ARTICLE{LozanoPerez1989,
  author={Lozano-Perez, T. and Jones, J.L. and Mazer, E. and O'Donnell, P.A.},
  journal={Computer}, 
  title={Task-level planning of pick-and-place robot motions}, 
  year={1989},
  volume={22},
  number={3},
  pages={21-29},}

@inproceedings{Fu2010, author = {Fu, Hongbo and Cohen-Or, Daniel and Dror, Gideon and Sheffer, Alla}, title = {Upright Orientation of Man-Made Objects}, year = {2008}, isbn = {9781450301121}, publisher = {Association for Computing Machinery}, booktitle = {ACM SIGGRAPH 2008 Papers}, articleno = {42},}

@INPROCEEDINGS{Saxena2009,
  author={Saxena, Ashutosh and Driemeyer, Justin and Ng, Andrew Y.},
  booktitle={IEEE ICRA}, 
  title={Learning 3-D object orientation from images}, 
  year={2009},
  pages={794-800},}

@INPROCEEDINGS{Holladay2013,
  author={Holladay, Anne and Barry, Jennifer and Kaelbling, Leslie Pack and Lozano-Pérez, Tomás},
  booktitle={2013 IEEE International Conference on Robotics and Automation}, 
  title={Object placement as inverse motion planning}, 
  year={2013},}

@article{zeng2020transporter,
    title={Transporter Networks: Rearranging the Visual World for Robotic Manipulation},
    author={Zeng, Andy and Florence, Pete and Tompson, Jonathan and Welker, Stefan and Chien, Jonathan and Attarian, Maria and Armstrong, Travis and Krasin, Ivan and Duong, Dan and Sindhwani, Vikas and Lee, Johnny},
    journal={CoRL},
    year={2020}
}

@article{YenChen2022NeRFSupervisionLD,
  title={NeRF-Supervision: Learning Dense Object Descriptors from Neural Radiance Fields},
  author={Lin Yen-Chen and Peter R. Florence and Jonathan T. Barron and Tsung-Yi Lin and Alberto Rodriguez and Phillip Isola},
  journal={IEEE ICRA},
  year={2022},
}

@INPROCEEDINGS{Bicchi2000,
  author={Bicchi, A. and Kumar, V.},
  booktitle={IEEE International Conference on Robotics and Automation.}, 
  title={Robotic grasping and contact: a review}, 
  year={2000},
  volume={1},
  pages={348-353 vol.1},}

@article{Salisbury1982,
author = {J. Kenneth Salisbury and John J. Craig},
title ={Articulated Hands: Force Control and Kinematic Issues},
journal = {IJRR},
volume = {1},
number = {1},
pages = {4-17},
year = {1982},
}

@ARTICLE{Miller2004,
  author={Miller, A.T. and Allen, P.K.},
  journal={IEEE Robotics \& Automation Magazine}, 
  title={Graspit! A versatile simulator for robotic grasping}, 
  year={2004},
  volume={11},
  number={4},
  pages={110-122},}

@article{sundermeyer2021contact,
  title={Contact-GraspNet: Efficient 6-DoF Grasp Generation in Cluttered Scenes},
  author={Sundermeyer, Martin and Mousavian, Arsalan and Triebel, Rudolph and Fox, Dieter},
  booktitle={ICRA},
  year={2021}
}

@inproceedings{mahler17dexnet2,
author = {Mahler, Jeffrey and Liang, Jacky and Niyaz, Sherdil and Laskey, Michael and Doan, Richard and Liu, Xinyu and Aparicio, Juan and Goldberg, Ken},
year = {2017},
month = {07},
pages = {},
title = {Dex-Net 2.0: Deep Learning to Plan Robust Grasps with Synthetic Point Clouds and Analytic Grasp Metrics},
  booktitle={RSS},
}

@INPROCEEDINGS{Pinto16grasp,
  author={L. {Pinto} and A. {Gupta}},
  booktitle={IEEE ICRA}, 
  title={Supersizing self-supervision: Learning to grasp from 50K tries and 700 robot hours}, 
  year={2016},
  pages={3406-3413},}

@article{mueller2022instant,
    author = {Thomas M\"uller and Alex Evans and Christoph Schied and Alexander Keller},
    title = {Instant Neural Graphics Primitives with a Multiresolution Hash Encoding},
    journal = {ACM Trans. Graph.},
    issue_date = {July 2022},
    volume = {41},
    number = {4},
    month = jul,
    year = {2022},
    pages = {102:1--102:15},
    articleno = {102},
    numpages = {15},
    publisher = {ACM},
    address = {New York, NY, USA}
}

@inproceedings{Mildenhall2020NeRFRS,
  title={NeRF: Representing Scenes as Neural Radiance Fields for View Synthesis},
  author={Ben Mildenhall and Pratul P. Srinivasan and Matthew Tancik and Jonathan T. Barron and Ravi Ramamoorthi and Ren Ng},
  booktitle={ECCV},
  year={2020}
}

@inproceedings{IchnowskiAvigal2021DexNeRF,
  title={{Dex-NeRF}: Using a Neural Radiance field to Grasp Transparent Objects},
  author={Ichnowski*, Jeffrey and Avigal*, Yahav and Kerr, Justin and Goldberg, Ken},
  booktitle={CoRL},
  year={2020}
}

@article{Pumarola2021DNeRFNR,
  title={D-NeRF: Neural Radiance Fields for Dynamic Scenes},
  author={Albert Pumarola and Enric Corona and Gerard Pons-Moll and Francesc Moreno-Noguer},
  journal={IEEE CVPR},
  year={2021},
  pages={10313-10322}
}

@article{Deng2021DepthsupervisedNF,
  title={Depth-supervised NeRF: Fewer Views and Faster Training for Free},
  author={Kangle Deng and Andrew Liu and Junyan Zhu and Deva Ramanan},
  journal={IEEE CVPR},
  year={2022},
}

@inproceedings{Wada2022,
title={{ReorientBot}: Learning Object Reorientation for Specific-Posed Placement},
author={Kentaro Wada and Stephen James and Andrew J. Davison},
booktitle={IEEE ICRA},
year={2022},
}

@inproceedings{ChavanDafle2022,
  author    = {Nikhil Chavan Dafle and
               Sergiy Popovych and
               Shubham Agrawal and
               Daniel D. Lee and
               Volkan Isler},
  title     = {Simultaneous Object Reconstruction and Grasp Prediction using a Camera-centric Object Shell Representation},
  booktitle   = {IEEE  IROS},
  year      = {2022},}

@INPROCEEDINGS{Schonberger2016colmap,  
author={Schönberger, Johannes L. and Frahm, Jan-Michael},  booktitle={IEEE CVPR},   title={Structure-from-Motion Revisited},   year={2016},  volume={},  number={},  pages={4104-4113}}

@article{Yu2021pixelNeRFNR,
  title={pixelNeRF: Neural Radiance Fields from One or Few Images},
  author={Alex Yu and Vickie Ye and Matthew Tancik and Angjoo Kanazawa},
  journal={IEEE CVPR},
  year={2021},
  pages={4576-4585}
}

@article{Song2020GraspingIT,
  title={Grasping in the Wild: Learning 6DoF Closed-Loop Grasping From Low-Cost Demonstrations},
  author={Shuran Song and Andy Zeng and Johnny Lee and Thomas A. Funkhouser},
  journal={IEEE Robotics and Automation Letters},
  year={2020},
  volume={5},
  pages={4978-4985}
}

@article{Pas2017GraspPD,
  title={Grasp Pose Detection in Point Clouds},
  author={Andreas ten Pas and Marcus Gualtieri and Kate Saenko and Robert W. Platt},
  journal={IJRR},
  year={2017},
  volume={36},
  pages={1455 - 1473}
}

@article{Roa14graspq,
author = {Roa, Maximo A. and Suarez, Raul},
year = {2014},
month = {07},
pages = {65-88},
title = {Grasp Quality Measures: Review and Performance},
volume = {38},
journal = {Autonomous Robots},
}

@INPROCEEDINGS{Wang2019,
  author = {Wang, Lirui and Xiang, Yu and Fox, Dieter},
  title = {Manipulation Trajectory Optimization with Online Grasp Synthesis and Selection},
  booktitle={RSS},
  year = {2019},
  }
